# Vector logic allows counterfactual virtualization by The Square Root of NOT

Eduardo Mizraji

Group of Cognitive Systems Modeling,
Biophysics and Systems Biology Section,
Facultad de Ciencias, Universidad de la República
Montevideo, Uruguay

## Abstract

In this work we investigate the representation of counterfactual conditionals using the vector logic, a matrix-vectors formalism for logical functions and truth values. Inside this formalism, the counterfactuals can be transformed in complex matrices preprocessing an implication matrix with one of the square roots of NOT, a complex matrix. This mathematical approach puts in evidence the virtual character of the counterfactuals. This happens because this representation produces a valuation of a counterfactual that is the superposition of the two opposite truth values weighted, respectively, by two complex conjugated coefficients. This result shows that this procedure gives an uncertain evaluation projected on the complex domain. After this basic representation, the judgment of the plausibility of a given counterfactual allows us to shift the decision towards an acceptance or a refusal. This shift is the result of applying for a second time one of the two square roots of NOT.

**Keywords**: Vector logic; Conditionals; Counterfactuals; Square Roots of NOT.

**Address:**

Dr. Eduardo Mizraji
Sección Biofísica y Biología de Sistemas,
Facultad de Ciencias, UdelaR
Iguá 4225, Montevideo 11400, Uruguay
e-mails: 1) emizraji@gmail.com, 2) mizraj@fcien.edu.uy
Phone-Fax: +598 25258629

# 1. Introduction

The comprehension and further formalization of the conditionals have a long and complex history, full of controversies, famously described in an article of Lukasiewicz [20]. Plausibly, material implication finally was relatively accepted due to its advantages to formalize deductive processes and demonstration rules. In fact, Lukasiewicz built the first version of his three-valued logic with postulates based on the implication [19]. A well-known alternative to material implication was the strict implication proposed by C.I. Lewis [17], a modal representation for the conditionals.

A topic in which the theory of the conditionals becomes important is the search of a logical formalization of counterfactuals, as was shown in the works of D. Lewis [18], Ginsberg [10] and Rescher [30] among many others. In this case, all the problems associated with the formalization of the conditionals become acute because counterfactuals add the no minor trouble of referring to facts or propositions placed out of the reality. This "virtuality" of counterfactual propositions produced an interesting connection of this problem with the logical theory of "possible worlds" [18].

In the present work, we explore the use of an operator emerged from the theory of quantum computing, the square root of NOT (symbolically, $\sqrt{\text{NOT}}$ or $\sqrt{\neg}$ ) to represent the virtual structure of conditional counterfactuals. In his famous article Hayes [13] confronts two logical gates that act on quantum logical values called qbits (vectors of dimension 2, similar to the ones we will illustrate in the next section). One of them is an operator that he calls "coin flip" (CoinFlip) and the other is the "quantum coin flip" (Q-CoinFlip). CoinFlip acts according to the classical probability calculus, and when the logical value "Input" is passed through two coin flip in series, (CoinFlip)*(CoinFlip), the result is uncertain according to the prediction of the probability calculations. But when "Input" enters a series of two quantum operators (Q-CoinFlip)*(Q-CoinFlip), the result is NOT("Input"). This leads to interpreting the Q-CoinFlip as $\sqrt{\text{NOT}}$. On the other hand, Deutsch et al. [7] show an optical device such that a pair of them, by exploiting quantum interference, also manage to operate as a negation, so that each is interpretable as $\sqrt{\text{NOT}}$. These authors show that the mathematics that describes the $\sqrt{\text{NOT}}$ involve complex variables. It is interesting to note that in their article the authors write *"Hence, reassured by the physical experiments that corroborate this theory, logicians are now entitled to propose a new logical operation $\sqrt{\text{NOT}}$. Why? Because a faithful physical model for it exists in nature!"* [7, pag. 271].

In our approach we translate the logic operations into the formalism of vector logic, with truth values represented by vectors and the logical functions by matrices [19, 20]. Consequently, linear algebra will be the basic language used along the present work. Vector logic was inspired by neurocomputational models of reasoning [18-21]. This is a field that led to the broad domain called "vector symbolic architectures" that incorporates an extensive series of cognitive models used in specific structures and procedures [1, 2, 15, 16, 21, 29, 32]. The use of tensor contextualization (in some cases with the Kronecker product format), fundamental in the present work, was initially presented in [22, 31].

The formalism that we present here, produces a very natural fuzzy logic when the logical truth values are linear combinations of the basic "true" and "false" vectors [23, 24, 25]. In addition, this linear algebra approach was relevant in the development of fuzzy logics based on complex vectors [8, 34].

We organize of the present work as follows. First, we present a brief review of some of the matrix operators involved in the representation of monadic connectives, including the square root of NOT, and the dyadic connectives implication, disjunction and conjunction. Then, we show a way to describe, inside this operator formalism, some subtle aspects of counterfactuals. Finally, we discuss the potential extensions of this approach.

## 2. A brief review of vector logic

Let $\tau = \{s, n : s, n \in \mathbb{R}^Q\}$ be a binary set of truth values. In this set $\tau$ we have the vector version of the classical truth values; they are the two orthonormal Q-dimensional column vectors "true", s, and "false", n. These vector truth values are arbitrary; we only ask for orthonormality. These truth values result from the mapping of a proposition Prop(u) on a vector truth value $u \in \tau$. The structure of s and n, as well as their dimension Q, are designed according to the nature of the problem. In this format, and adopting the conventional matrix expressions, the identity logical operator is a matrix $I \in \mathbb{R}^{Q \times Q}$ given by

$$I = nn^T + ss^T \tag{1}$$

and the negation $N \in \mathbb{R}^{Q \times Q}$ is given by

$$N = sn^T + ns^T. \tag{2}$$

The superindex T means transposition. Remark that $N^2 = I$.

In [25] a general expression of the square root of N, the logical gate initially discovered in the context of quantum computing [7, 13], is presented. Here we adopt the following representation of the two roots $\sqrt{N}$:

$$A = \left(\sqrt{N}\right)_1 = \tfrac{1}{2}(1+i)I + \tfrac{1}{2}(1-i)N, \tag{3}$$

$$B = \left(\sqrt{N}\right)_2 = \tfrac{1}{2}(1-i)I + \tfrac{1}{2}(1+i)N, \tag{4}$$

with $i = \sqrt{-1}$.

These matrices A and B have the following remarkable properties:

$$A^2 = N \; ; \quad B^2 = N \; ; \tag{5}$$

$$AB = BA = I. \tag{6}$$

Another interesting point is the analogy with the two squares roots of -1. The positive root $+(\sqrt{-1})$ corresponds to

$$\left(\sqrt{N}\right)_1 = IA,$$

and the negative root $-(\sqrt{-1})$ corresponds to

$$\left(\sqrt{N}\right)_2 = NA \ ;$$

as a consequence, $NA = B$.

In what follows, after the formal definitions of the logical operators, we will illustrate them using two sets of 2-dimensional truth values:

Set 1: $s = \begin{bmatrix} 1 \\ 0 \end{bmatrix}$, $n = \begin{bmatrix} 0 \\ 1 \end{bmatrix}$ ; Set 2: $s = \dfrac{1}{\sqrt{2}}\begin{bmatrix} 1 \\ 1 \end{bmatrix}$, $n = \dfrac{1}{\sqrt{2}}\begin{bmatrix} 1 \\ -1 \end{bmatrix}$.

*Example 1.* Monadic operators for Set 1 and Set 2

Set 1: $I = \begin{bmatrix} 1 & 0 \\ 0 & 1 \end{bmatrix}$, $N = \begin{bmatrix} 0 & 1 \\ 1 & 0 \end{bmatrix}$, $A = \begin{bmatrix} \frac{1}{2}(1+i) & \frac{1}{2}(1-i) \\ \frac{1}{2}(1-i) & \frac{1}{2}(1+i) \end{bmatrix}$, $B = \begin{bmatrix} \frac{1}{2}(1-i) & \frac{1}{2}(1+i) \\ \frac{1}{2}(1+i) & \frac{1}{2}(1-i) \end{bmatrix}$

Set 2: $I = \begin{bmatrix} 1 & 0 \\ 0 & 1 \end{bmatrix}$, $N = \begin{bmatrix} 1 & 0 \\ 0 & -1 \end{bmatrix}$, $A = \begin{bmatrix} 1 & 0 \\ 0 & i \end{bmatrix}$, $B = \begin{bmatrix} 1 & 0 \\ 0 & -i \end{bmatrix}$

Now we define the Kronecker product, an important operation for the following arguments. Being U and V two matrices, no matter what its dimensions, the Kronecker product $U \otimes V$ is defined as follows [12]:

$$U \otimes V = \begin{bmatrix} u_{ij} V \end{bmatrix} \ .$$

*Example 2.* Numerical Kronecker product (the subindex describes the matrix dimension)

$$U = \begin{bmatrix} 1 & 0 \\ 2 & -1 \end{bmatrix}_{2\times 2} ; \ V = \begin{bmatrix} 1 & -1 & 4 \\ 3 & 1 & 0 \end{bmatrix}_{2\times 3} ; \ U \otimes V = \begin{bmatrix} 1V & 0V \\ 2V & -1V \end{bmatrix} = \begin{bmatrix} 1 & -1 & 4 & 0 & 0 & 0 \\ 3 & 1 & 0 & 0 & 0 & 0 \\ 2 & -2 & 8 & -1 & 1 & -4 \\ 6 & 2 & 0 & -3 & -1 & 0 \end{bmatrix}_{4\times 6}$$

We only show here two of the properties of the Kronecker product (for details, see [12]):

(a) $(U \otimes V)^T = U^T \otimes V^T$
(b) $(U_1 \otimes V_1)(U_2 \otimes V_2) = (U_1 U_2) \otimes (V_1 V_2)$ .

In (a) the matrices have no dimensional restrictions; in (b) the products $U_1 U_2$ and $V_1 V_2$ must be possible. In the case of four column vectors $a, b, c, d \in \mathbb{R}^Q$, we have

$$(a \otimes b)^T (c \otimes d) = (a^T c) \otimes (b^T d) = \langle a, b \rangle \langle c, d \rangle \ ,$$

with the symbol $\langle\,,\,\rangle$ indicating the scalar product. These two scalar products are the mathematical filters that allow us to represent the dyadic logical operators in terms of linear algebra [24, 25, 26].

Now we return to logic. The basic table that defines the material implication $p \to q$, the disjunction $p \vee q$, and the conjunction $p \wedge q$, defined for the symbolic truth values "true", t, and "false", f, is the following:

| p | q | $p \to q$ | $p \vee q$ | $p \wedge q$ |
|---|---|-----------|------------|--------------|
| t | t | t | t | t |
| t | f | f | t | f |
| f | t | t | t | f |
| f | f | t | f | f |

We remind a basic equivalence that defines the material implication in terms of the disjunction: $p \to q \equiv \neg p \vee q$.

Now, we present the matrix operators that implement the material implication $p \to q$, the disjunction $p \vee q$ and the conjunction $p \wedge q$ over the vector truth values s and n. First, consider the following matrix:

$$H = [(s \otimes s) \ (s \otimes n) \ (n \otimes s) \ (n \otimes n)] \qquad (7)$$

with $u \otimes v$ $(u, v \in \tau)$ being the Kronecker product. We can represent the matrices implication L, disjunction D and conjunction C, respectively, by the following compact equations (8):

$$L = [s\ n\ s\ s]H^T, \quad D = [s\ s\ s\ n]H^T, \quad C = [s\ n\ n\ n]H^T \qquad (8)$$

being [s n s s], [s s s n] and [s n n n] partitioned matrices in $\mathbb{R}^{Q \times 4}$. Remark that $H^T \in \mathbb{R}^{4 \times Q^2}$, hence, $L, D, C \in \mathbb{R}^{Q \times Q^2}$. It is interesting to note the similarity of these matrix representations with the definitions of implication, disjunction and conjunction displayed in Table 5.101 of the Wittgenstein's *Tractatus* [33]. For instance, in Table 5.101 of his *Tractatus*, naming true T and false F, Wittgenstein defines the implication, the disjunction and the conjunction as follows:

"(TFTT)(p,q) in words If q then p [q ⊃ p]"
"(TTTF)(p,q) in words p or q [p ∨ q]"
"(TFFF)(p,q) in words p and q [p.q]"

In these definitions, Wittgenstein symbolizes by (p,q) the different truth value pairs (TT), (TF), (FT), (FF).

In an extended version, matrices (8) can be written as

$$L = s(s \otimes s)^T + n(s \otimes n)^T + s(n \otimes s)^T + s(n \otimes n)^T, \qquad (9)$$
$$D = s(s \otimes s)^T + s(s \otimes n)^T + s(n \otimes s)^T + n(n \otimes n)^T, \qquad (10)$$
$$C = s(s \otimes s)^T + n(s \otimes n)^T + n(n \otimes s)^T + n(n \otimes n)^T. \qquad (11)$$

*Example 3*: The basic matrix logic operators defined for Set 1 and Set 2.

Set 1: $L = \begin{bmatrix} 1 & 0 & 1 & 1 \\ 0 & 1 & 0 & 0 \end{bmatrix}, \; D = \begin{bmatrix} 1 & 1 & 1 & 0 \\ 0 & 0 & 0 & 1 \end{bmatrix}, \; C = \begin{bmatrix} 1 & 0 & 0 & 0 \\ 0 & 1 & 1 & 1 \end{bmatrix}$

Set 2: $L = \dfrac{1}{\sqrt{2}} \begin{bmatrix} 2 & 0 & 0 & 0 \\ 1 & 1 & -1 & 1 \end{bmatrix}, \; D = \dfrac{1}{\sqrt{2}} \begin{bmatrix} 2 & 0 & 0 & 0 \\ 1 & 1 & 1 & -1 \end{bmatrix}, \; C = \dfrac{1}{\sqrt{2}} \begin{bmatrix} 2 & 0 & 0 & 0 \\ -1 & 1 & 1 & 1 \end{bmatrix}$

The computing capacity of these operators is based on the properties of the Kronecker product. As a consequence, we have for the matrix implication, disjunction and conjunction, the following results:

$L(s \otimes s) = L(n \otimes s) = L(n \otimes n) = s$ and $L(s \otimes n) = n$;
$D(s \otimes s) = D(s \otimes n) = D(n \otimes s) = s$ and $D(n \otimes n) = n$;
$C(s \otimes s) = s$; and $C(s \otimes n) = C(n \otimes s) = C(n \otimes n) = n$.

These matrices lead to a capital finding that distinguishes this formalism from classical formalism of the standard logic. We show it with an example. Imagine we want to express in the vector-matrix format the classical logic equivalence between implication and disjunction $p \to q \equiv \neg p \vee q$. So, assuming that u and v are some of the values s and n ($u, v \in \tau$), we write

$$L(u \otimes v) = D(Nu \otimes v).$$

But, due to the property (b) of the Kronecker product, we can factorize this equation as follows:

$$L(u \otimes v) = D(Nu \otimes Iv) = D(N \otimes I)(u \otimes v).$$

It is well known that matrix factors sometimes cannot be treated and simplified as in simple algebraic equations between numbers. But in this case, it can be directly proved from equations (1), (2), (9) and (10) that the following equality between operators is satisfied:

$$L = D(N \otimes I).$$

Consequently, in this formalism, some important logical tautologies become equalities between matrix operators, independent of vector variables. This is the case, for instance, for the De Morgan Laws that produce the matrix equalities $C = ND(N \otimes N)$ and $D = NC(N \otimes N)$ [23].

Another important point of this matrix-vector formalism, is that these binary (or Boolean) matrices, defined over the two basic vectors s and n, can process fuzzy truth vectors, generated as a weighted linear combination of s and n. In this way, we get a fuzzy logic with fuzziness imposed by the inputs and no by the operators (for details see [23, 24, 26]). In this context, it is interesting to mention the contribution of Dalla Chiara et al. [6] that generates a fuzzy quantum logic replacing the notion of qubit with that of qudit, a unit vector of many dimensions that obtains degrees of fuzziness displacing 1 between the zeros from the "true" vector and the "not" vector.

## 3. Counterfactuals

The central aspect of counterfactuals is that they refer to non-existent situations and establish implicative rapports between events contradicted by the reality. This interaction with the reality put in front the issue of counterfactuals belong both to the field of logic and that of physics, including in "physics" all the events that happen in our real world. This double face of the counterfactuals, formal and physical, explains the fact that theorizing about counterfactuals becomes mixed with theorizing about causality or the sense of time [14, 18, 28]. By the moment we are going to omit these important points and to concentrate in trying to state a formalism capable to deal with the fictional character of counterfactuals. In fact, our intention is to introduce this new formalism to the wide and diverse community of researchers that approaches the subtleties of counterfactuals from so many angles.

Ginsberg [10] proposed the following definition: *"A counterfactual is a statement such as, 'if p, then q,' where the premise p is either know or expected to be false"*, then he adds that *"falsehood implies anything"*. This author gives us two examples of counterfactuals with a different qualitative level of plausibility. On the one hand, we can state *"If the electricity hadn't failed, dinner would have been ready on time"*; on the other, we can state *"If the electricity hadn't failed, pigs would fly"* [10]. Based on his notation, we can write the "truth table" (assuming that this expression is acceptable in this context) of a counterfactual conditional $p^* < q^*$ as follows:

| $p^*$ | $q^*$ | $p^* < q^*$ |
|---|---|---|
| t | t | t |
| t | f | f |
| f | t | ? |
| f | f | ? |

Here, $p^*$ and $q^*$ are the modified counterfactual propositions derived from a true conditional $p \to q$.

Rescher [30] analyzes the logical possibilities of the "what …. if" situations and organizes these possibilities of creating counterfactuals in two classes:

I. "If p (which is true) were false, then what?"
II "If p (which is false) were true, then what?"

In this work, we are going to explore a minimalist approach that focuses a restricted category of counterfactuals. First, we define the "factual conditional", FC, as an implication necessarily true. This assumption is based on the nature of our real world that shows factual conditionals, like "if we don't eat then we die", empirically true. The reality of this conditional can, in fact, be considered a definition of "true". Then, we are going to assume that the counterfactual implication associated to a particular FC, let us symbolize it by CF, is *a priori* true. Hence, the paradigm of this minimalist approach is that $p \to q$ "true" becomes $p^* < q^*$ "hypothetically true". The heuristic behind the last supposition is the following. In our cognitive life, we construct alternatives looking for plausible conjectural conditionals. These counterfactuals are created usually to investigate interesting organizational, economical, sociological, or technological strategic possibilities not existent yet. The counterfactual builders usually try to explore future possible real situations. For this reason, we can tentatively suppose an *a priori* hypothetical true for any counterfactual. In the following subsection a refined way to judge the value of this tentative valuation is described.

To simplify the notation, let us represent the propositions and the corresponding truth values with the same notation. The detailed formalism that establishes the correspondence between propositions and vector truth values is described in [25]. Using the matrix formalism, the link between factual and counterfactual conditionals looks as follows:

$$L(p \otimes q) = s \xrightarrow{\text{counterfactual}} L^c(p^* \otimes q^*) \triangleq s \qquad (13)$$

with $L^c$ being the counterfactual conditional, $p^*, q^* \in \tau$ and the symbol $\triangleq$ is used here to represent the conjectural *a priori* assumed truth value. This format excludes counterfactuals with $p^* = s$ and $q^* = n$.

The transformation represented in (13) symbolizes the conversion of a true conditional into a conjecturally true counterfactual. The symbol $L^c$ represents the algebraic matrix associated to the classic implication "if ... then", but the superindex c indicates that this operation is applied on the counterfactual propositions $p^*$ and $q^*$, that correspond to statements such as "if it had been 'X' then 'Y' would have resulted", or similar expressions.

### 3.1. Virtualization

Taking into account that counterfactuals don't belong to the real world, in order to transform these statements into virtual propositions, that we symbolize by CF(v), we can exploit the properties of the $\sqrt{\text{NOT}}$ and premultiply CF by the matrix A or the matrix B (in this stage, it is not relevant the selected root). If we select A, the results of this "virtualization" are the following:

(a) CF(v): $AL^c(s \otimes s) = As$

(b) CF(v): $AL^c(n \otimes s) = As$ $\qquad (15)$

(c) CF(v): $AL^c(n \otimes n) = As$

Note that $L \in \mathbb{R}^{Q \times Q^2}$ but $AL^c, BL^c \in \mathbb{C}^{Q \times Q^2}$. Particularly important is what happen with the truth values of these virtual conditionals. Remark that

$$As = \tfrac{1}{2}(1+i)s + \tfrac{1}{2}(1-i)n \ , \qquad (16)$$

$$Bs = \tfrac{1}{2}(1-i)s + \tfrac{1}{2}(1+i)n \ . \qquad (17)$$

This is a suggesting result because in this formalism the uncertainty of a counterfactual valuation is represented by the splitting, by the matrices A and B, of the truth value associated to the counterfactual conditional into the two superimposed complex truth values, $\tfrac{1}{2}(1 \pm i)s$, a complex version of "true", and $\tfrac{1}{2}(1 \mp i)n$, the complex version of "false". In the virtual world, both complex truth values are equally valid. Hence, under this representation, the counterfactual plausibility is uncertain.

### *3.2. Back to real*

Given a virtual conditional expressed by a counterfactual, $p^* < q^*$ derived from a real conditional $p \rightarrow q$, we need to explore its level of plausibility. The evaluation of the sources of plausibility have been soundly analyzed by many authors using different approaches and methods (e.g.: Goodman [11], Lewis [18], Horwich [14], Pearl [28], Bochman [5]). Here, our criteria of plausibility is double. On the one hand, we require that the hypothetical relation between the antecedent $p^*$ and consequent $q^*$ can happen in the real world. On the other hand, we need this hypothetical implication to be logically consistent. Let us assume that there are a set including the plausible factual evidence F and a set of logical laws C.

F is the set of facts that has the approval of human experience. A statement like "the national elections could have been won by John Doe" belongs to F. Instead, "the elephants could fly by waving their ears" does not belong to F.

C is the set of propositions that have logical consistency. For example, the statement "The double negation is equivalent to an affirmation" has logical consistency and belongs to C, instead "the addition of -4 to 231 equals 235" does not belong to C.

Now, given the counterfactual proposition $p^*$ and $q^*$, we are going to use the logical formula (18) to evaluate the plausibility of this counterfactual:

$$PL(p^*, q^*) \equiv \left[ \left[ (p^* \in F) \wedge (q^* \in F) \right] \wedge \left[ (p^* \in C) \wedge (q^* \in C) \right] \right] \ . \qquad (18)$$

Based on this formula we assume that a counterfactual is plausible if $PL(p^*, q^*) \equiv \text{true}$ and implausible if $PL(p^*, q^*) \equiv \text{false}$.

Returning to the examples of Ginsberg [10], if we consider the counterfactual "If the electricity hadn't failed, then dinner would have been ready on time" under the form $p^* < q^*$, we see that the antecedent and the consequent are both consistent with possible empirical facts F, and that they are not inconsistent with any formal logic law. Hence, the three conjunctions of eq. (18) are true and the final evaluation is true. Consequently, we can consider plausible this counterfactual. In the case of "If the electricity hadn't failed, then pigs would fly", is clear that the second part of this counterfactual, q*, is not real. Consequently, we can assume that $q* \notin F$, and being the first conjunction false, the whole expression (18) is false. And we can assume implausible this second counterfactual. Remark that the factual implication associated with the previous counterfactuals can be expressed as follows (1) "If the electricity fails, then dinner cannot be ready on time", and (2) "If the electricity fails, then pigs cannot fly", In both cases, the associated factual conditionals FC have the same structure: $L(p \otimes q) = L(s \otimes s)$. This is because in both cases the propositions p and q are factually plausible. And the associated counterfactuals result from the negation of both p and q given $L^c(p* \otimes q*) = L^c(n \otimes n)$.

A nice aspect of the formal virtualization generated by the square roots of NOT, is that we have a simple way to turn the virtual (and neutral) evaluations given in equations (16-17) into real evaluations established according some plausibility criteria. The key is to premultiply the virtual expressions by one of the square roots of NOT, A or B, according to the structure of the virtual counterfactual and the decision about plausibility or implausibility.

To define a way to project the level of plausibility into the matrix formalism, we proceed as follows. First we notice that matrices A and B are complex conjugate: $A* = B$ and conversely $B* = A$. Now we can define a matrix plausibility $mpl(CF(v))$ function for a given counterfactual CF by:

$$mpl(CF(v)) = \begin{cases} X*[XL(p*\otimes q*)] & \text{if } PL(p*,q*) \equiv \text{true} \\ X[XL(p*\otimes q*)] & \text{if } PL(p*,q*) \equiv \text{false} \end{cases} \quad (19)$$

with $X \in \{A, B\}$.

We illustrate these evaluations presenting an example of counterfactuals having the same syntactic structure but different plausibility. We begin quoting the insightful comment of Emerson [9] about the nature of genius, included in his essay "Shakespeare, or the Poet". Emerson wrote *"There is no choice to genius. (…) he finds himself in the river of the thoughts and events, forced onward by the ideas and necessities of his contemporaries. He stands where all the eyes of men look one way, and their hands all point in the direction in which he should go."* Let us assume that Emerson is right and that any genius fulfill necessities of its own epoch. However, there is an important difference between artistic genius and scientific genius: the creation of the genial artist would be impossible to recreate if he had not existed, but the creation of a genial scientist would be very likely to be recreated if he had not existed. With this in mind, we present the following two counterfactuals referred to persons considered genius of their times:

(a) If Jorge Luis Borges had not been born, the story "Death and the Compass" would not have been written.

(b) If Thomas Willis had not been born, the cerebral arterial circle would not have been discovered.

Both can be represented by a conditional $p \to q$ and a counterfactual resulting from the negation of the antecedent and the consequent, $p^* < q^*$, with $p^* = \neg p$ and $q^* = \neg q$. We can associate to these situations the corresponding factual conditional FC, the associated counterfactual CF and virtual counterfactual CF(v):

FC: $L(s \otimes s) = s$,

CF: $L^c(n \otimes n) = s$,

CF(v): $AL^c(n \otimes n) = As = \frac{1}{2}(1+i)s + \frac{1}{2}(1-i)n$.

But these counterfactuals with the same syntactical structure, have different plausibility. The first, is almost sure. The factual probability that other person wrote the same story, in Spanish, with exactly the same words is almost zero, and consequently, the probability of q* is almost one. Consequently, our criteria of plausibility (18-19) give true and we can render true our counterfactual (a). To do this, we need to premultiply the virtual counterfactual by matrix B, resulting

$mpl(CF(v)) = BAL^c(n \otimes n) = BAs = s$.

In the second case the situation is different. In the time of Thomas Willis (1621-1675) the anatomical research was intense and it is almost sure that the beautiful arterial structure, named frequently "Willis polygon", would have been equally discovered if Willis had not existed. In this case, the probability of $q^* = \neg q$ is near zero and according to our equations (18-19) the counterfactual is false. Consequently, we must negate our virtual counterfactual (b). In terms of the matrix format, we should have

$mpl(CF(v)) = AAL^c(n \otimes n) = AAs = n$.

It is particularly important to emphasize that, in this framework, the decision on counterfactual plausibility depends on aspects that are outside its formal structure.

## 4. Discussion

The roots of this work were the investigations of how to represent the operations of standard logic using a matrix-vector formalism. Originally, these investigations were motivated by the confirmation that the cognitive data inside the brain structures, map on very huge sets of electrochemical potentials, very naturally represented by large vectors. The associative memories that store and link these cognitive data can be represented by large dimensional matrices [1, 16]. In the neural framework, it is natural to assume that decisions about the truth or falsity of propositions can be represented by vectors. The mapping of truth values on large dimensional vectors produced, on the theoretical side, unexpected results. We are going to indicate some of them: 1) the vector truth values induce the construction of matrix operators that represent associative memories that encode the basic logical gates (negation, conjunction, disjunction, material implication, etc.); 2) the binary (Boolean) matrix operators, i.e. matrices based on two truth values, are capable to accept fuzzy vectors as inputs, represented by a probabilistic addition of true and false vectors; in front of these inputs, binary matrices produce fuzzy logics, with the fuzziness imposed by the inputs and not by the operators; 3) some important tautologies (e.g. the De Morgan Laws) become identities between the matrix operators involved, with no participation of the logical variables; 4) the modal operators possibility and necessity are easily constructed from these standard logic operators [25, 26]. However, if we want to represent counterfactual conditionals within this formalism, using the implication matrix L as our formal representation of the conditional operator, we encounter the serious difficulty that the semantics reside outside the formalism. How to resolve that two implication matrices, of identical structure, are assigned for as different objectives as a classical implication and a counterfactual conditional? The solution proposed here is to apply two actions on the counterfactual implication (whose operator is the matrix basic implication, but which we mark here as $L^c$). The first action is the premultiplication of the matrix counterfactual by one of the square roots of NOT. This action projects the decision into two virtual "possible worlds", each containing one of the two contradictory vector decisions in the complex field. The second action moves the problem out of the domain of logic, and evaluates the plausibility of the counterfactual proposition according to its logical and physical consistency, relative to the context of the situation. This second action decides, depending on the plausibility or implausibility of the proposition, a way of biasing the decision towards an approval or a rejection, respectively represented by the true or false real vectors, as expressed in equation (19). Finally, let us point out that with the advent of research on quantum computing, another remarkably vigorous development path appeared for vector-matrix representations of logical operations (3, 6, 35). Therefore, finding ways to explore the potentialities of counterfactual propositions that include novel proposals or alternative procedures, can become an important computational goal.

Of course, this work does not pretend to describe the way cognitive neural systems try to solve the decisions concerning counterfactuals. It is clear that our minds can deal with counterfactuals, and that they are constantly present in our social life. The actual knowledge about brain function shows that the neural processing of complex information is carried out by large neural networks with the ability to process in parallel a large amount of complex information [1, 2, 16, 21, 27, 32]. These large neural systems, that in fact model accurately a good deal of our cognitive functions, are capable to map into a memory a fictional counterfactual and to sustain the two contradictory options of the eventual counterfactual decision. For the modeling of these neural abilities, the complex quantum operators do not appear as a physical necessity. Hence, in the present work, the square roots of NOT are used as an interesting formal device, not as a cognitive model. However, there is now a large amount of neural research that uses the statistical properties of quantum operators to analyze results emerged from experiments in the domain of cognitive sciences. This is territory of intensive research that exploits formalisms born in quantum physics (for a comprehensive introduction to these approaches, see [3, 4]).

The research on counterfactuals shows nowadays an extraordinary expansion. As stated earlier, the purpose of this work is not to delve into the vast theoretical corpus about counterfactuals but to add another instrument to the large repertoire of theoretical methods used in this research area.

Finally, a comment about the introduction of the square roots of NOT in basic logic. It is well known that the introduction of complex numbers in many branches of science turned out to be a beneficial procedure to enlarge the technical power of a theory and to obtain unified concepts. In this sense, a possible extension of this work is to study the representation of the logical connectives and the associated tautologies using a matrix format with the square roots on NOT acting as logical variables. As an example, we comment that we can express the disjunction as $D(X,Y) = D(XBs \otimes YAn)$, being the truth values s and n parameters, and the variables being $X,Y \in \{A,B\}$. As a consequence, we can express all the other logical operations taking into account that $NOR(X,Y) = ND(X,Y)$. If we think in logical circuits, the fact that the gates are physically implementable, including the $\sqrt{NOT}$ (see [7] and [13]), may add a further interest to the possibility to built logical circuits with the truth values acting as parameters and the gates as switching variables.

**Acknowledgments.** The author thanks to CSIC-UdelaR and ANII for partial financial support. He wishes to thank both reviewers for very helpful comments and insights.

# References

[1] J.A. Anderson. *An Introduction to Neural Networks*. MIT Press, Cambridge. MA., 1995.

[2] M.A. Arbib (Ed). *The Handbook of Brain Theory and Neural Networks*. MIT Press, Cambridge. MA. 1995.

[3] R. Blutner.. Questions and Answers in an Orthoalgebraic Approach. *Journal of Logic, Language and Information*, 21, 237-277, 2012

[4] R. Blutner and P. beim Graben. Quantum cognition and bounded rationality. *Synthese, 193,* 3239-3291, 2016

[5] A. Bochman. On laws and counterfactuals in causal reasoning, *Proceedings of the Sixteenth International Conference on Principles of Knowledge Representation and Reasoning*, 494-503, 2018.

[6] M.L. Dalla Chiara, R. Giuntini, G. Sergioli and R. Leporini. A many-valued approach to quantum computational logics. *Fuzzy Sets and Systems, 335,* 94-111, 2018

[7] D. Deutsch, A. Ekert and R. Lupacchini. Machines, logic and quantum physics. *The Bulletin of Symbolic Logic, 6,* 265-283, 2000.

[8] S. Dick. Towards complex fuzzy logic. *IEEE Transactions on Fuzzy Systems, 15,* 405–414, 2005.

[9] R.W. Emerson. *Representative Men: Seven Lectures*, Phillips, Sampson and Co, Boston, 1850.

[10] M.L. Ginsberg. Counterfactuals, *Artificial Intelligence, 30,* 35-79, 1986.

[11] N. Goodman. The problem of counterfactual conditionals, *The Journal of Philosophy, 44,* 113-128. 1947.

[12] A. Graham. *Kronecker Products and Matrix Calculus with Applications.* Ellis Horwood, Chichester, 1981.

[13] B. Hayes. The square root of NOT. *American Scientist, 83,* 304–308, 1995.

[14] P. Horwich. *Asymmetries in Time.* The MIT Press, Cambridge, MA., 1989.

[15] D. Kleyko1 , E. Osipov1 and R. W. Gayler. Recognizing permuted words with Vector Symbolic Architectures: A Cambridge test for machines. *Procedia Computer Science. 88,* 169–175, 2016

[16] T. Kohonen. *Associative Memory: A System-Theoretical Approach.* Springer-Verlag, New York, 1977.

[17] C.I. Lewis and C.H. Langford. *Symbolic Logic.* Dover, New York, 1959.

[18] D. Lewis. *Counterfactuals*, Blackwell, Oxford, 1973

[19] J. Lukasiewicz, On Three-Valued Logic [1920]. *in J. Lukasiewicz, Selected Works, L. Borkowski, ed.*, pp. 153–178. North-Holland, Amsterdam, 1980.

[20] J. Lukasiewicz. On the History of the Logic of Propositions [1934]. *in J. Lukasiewicz, Selected Works, L. Borkowski, ed.*, pp. 153–178. North-Holland, Amsterdam, 1980.

[21] J.L.McClelland, D.E. Rumelhart and the PDP Research Group. Parallel Distributed Processing. *Explorations in the Microstructure of Cognition.*(Psychological and Biological Models). The MIT Press, Cambridge, MS, 1986.

[22] E. Mizraji. Context-dependent associations in linear distributed memories. *Bull. Math. Biol.*, *51*, 195–205 1989.

[23] E. Mizraji. Vector logics: the matrix-vector representation of logical calculus, *Fuzzy Sets and Systems*, *50*, 179-185, 1992.

[24] E. Mizraji. The operators of vector logic, *Mathematical Logic Quarterly*, *42*, 27-40, 1996.

[25] E. Mizraji. Vector logic: a natural algebraic representation of the fundamental logical gates, *Journal of Logic and Computation*, *18*, 97-121, 2008.

[26] E. Mizraji. *El Álgebra Matricial de la Lógica*. Librería Linardi y Risso, Montevideo [Digital version posted in Research Gate (www.researchgate.net) and Academia (www.academia.edu)], 2019.

[27] E. Mizraji and J. Lin. Logic in a dynamic brain, *Bulletin of Mathematical Biology*, *71*, 373-379, 2011.

[28] J. Pearl. *Causality*. Cambridge University Press, Cambridge, 2000.

[29] T. Plate. Holographic reduced representations. Technical Report CRG-TR-91-1, Department of Computer Science, University of Toronto, 1991

[30] N. Rescher. *Conditionals*. The MIT Press, Cambridge, MA., 2007.

[31] P. Smolensky. Tensor product variable binding and the representation of symbolic structures in connectionist systems. *Artificial Intelligence. 46*, 159–216. 1990

[32] O. Sporns O. *Networks of the Brain*, The MIT Press, Cambridge, MA., 2010.

[33] L. Wittgenstein, L *Tractatus-Logico-Philosophicus* [1921], Routledge, London, 1974.

[34] O. Yazdanbakhsh and S. Dick. A systematic review of complex fuzzy sets and logic, *Fuzzy Sets and Systems, 338*, 1-22, 2018.

[35] Y. Younes, and I. Schmitt. On quantum implication. Quantum Machine Intelligence, *1(1-2)*, 53-63, 2019.